\documentclass{egpubl}
\usepackage{egsr2021_DLOnlyTrack}
 
\WsPaper           %

\usepackage[T1]{fontenc}
\usepackage{dfadobe}

\usepackage{booktabs}
\usepackage{siunitx}

\usepackage{amsmath}
\usepackage{amssymb}
\biberVersion
\BibtexOrBiblatex

\usepackage[backend=biber,bibstyle=EG,citestyle=alphabetic,backref=true]{biblatex} 
\electronicVersion
\PrintedOrElectronic

\ifpdf \usepackage[pdftex]{graphicx} \pdfcompresslevel=9
\else \usepackage[dvips]{graphicx} \fi

\usepackage{egweblnk}

\title[Single-image Full-body Human Relighting]{Single-image Full-body Human Relighting}

\author[Lagunas et al.]
{
    \parbox{\textwidth}{
        \centering Manuel Lagunas$^{1,2}$, Xin Sun$^{2}$, Jimei Yang$^{2}$, Ruben Villegas$^{2}$, 
        Jianming Zhang$^{2}$, Zhixin Shu$^{2}$, Belen Masia$^{1}$, and Diego Gutierrez$^{1}$
    }
    \\
    {\parbox{\textwidth}
                {\centering 
                $^1$Universidad de Zaragoza, I3A
                \\
                $^2$Adobe Research
        }
    }
}

\begin{document}

\maketitle

\newcommand{\dwi}[0]{\text{d}\omega_i}
\newcommand{\wi}[0]{\omega_i}

\begin{abstract}
    We present a single-image data-driven method to automatically relight images with full-body humans in them. Our framework is based on a realistic scene decomposition leveraging precomputed radiance transfer (PRT) and spherical harmonics (SH) lighting. In contrast to previous work, we lift the assumptions on Lambertian materials and explicitly model diffuse and specular reflectance in our data. Moreover, we introduce an additional light-dependent residual term that accounts for errors in the PRT-based image reconstruction. We propose a new deep learning architecture, tailored to the decomposition performed in PRT, that is trained using a combination of L1, logarithmic, and rendering losses. Our model outperforms the state of the art for full-body human relighting both with synthetic images and photographs.


\end{abstract}

\section{Introduction}
    The growth in mobile computing, together with the increasing demand for visual social media has led to a tremendous rise in the popularity of consumer digital photography. In full-body photographs lighting plays an important role in transmitting the desired appearance of the subject, and changes in the illumination can lead to drastically different renditions. However, these photographs usually lack controlled illumination conditions. 
    
\begin{figure}
        \centering
        \includegraphics[width=0.95\linewidth]{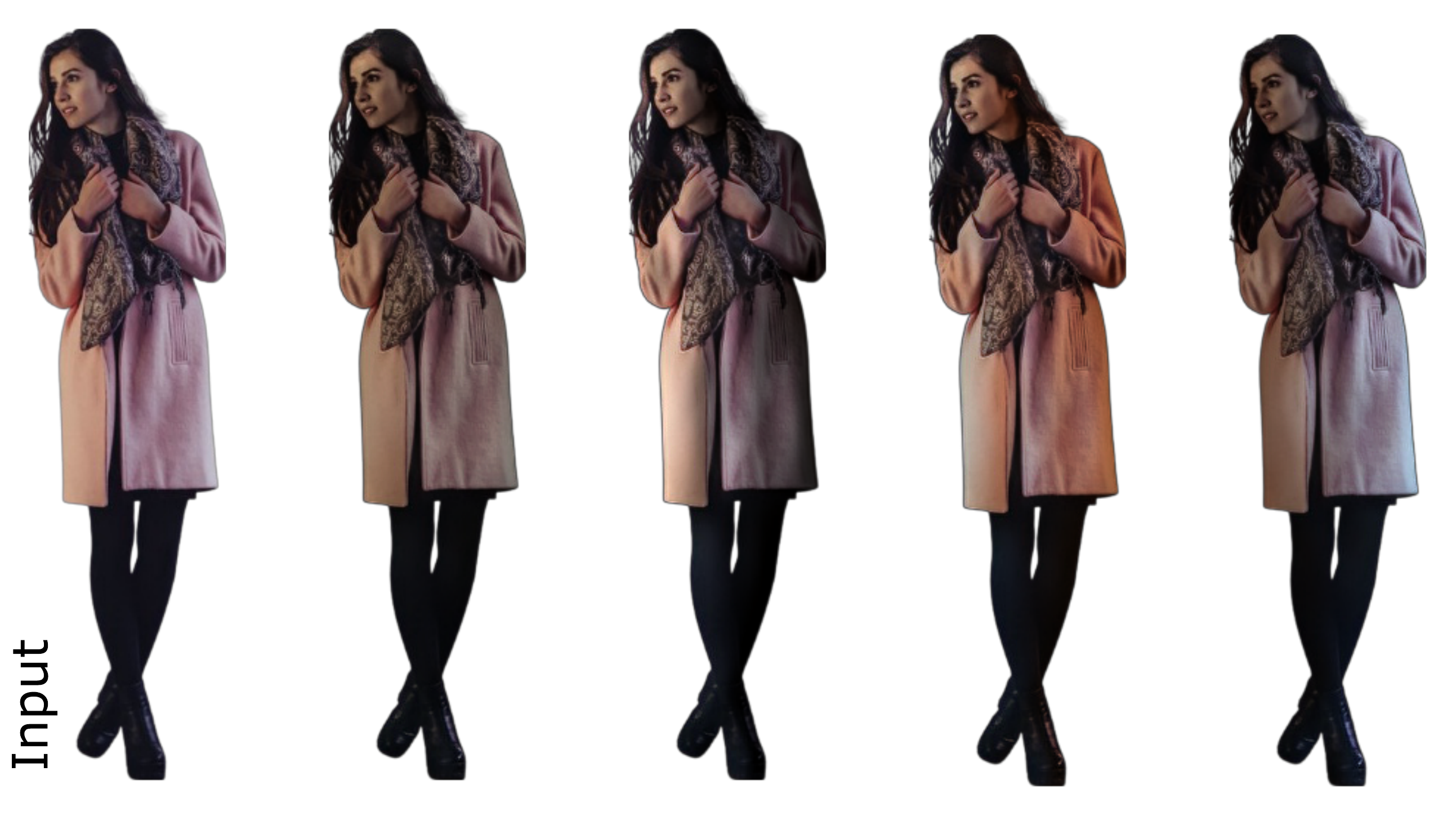}
        \caption{Relighted results given a single image as input for different illumination maps. Please refer to Figure \ref{fig:results_photos_relight} for more details about the reconstructions. }
        \label{fig:teaser}
    \end{figure}

    We present a single-image relighting method that acts as a post-processing step, allowing a casual user to plausibly change and manipulate the illumination on a subject in a photograph.
    Human relighting usually benefits from multiple images as input, and requires solving an inverse rendering problem; in the general case, illumination information needs to be disambiguated from geometry and material appearance, based on simple pixel values. This is a well-studied but ill-posed problem, for which no definite solution exists. This paper takes a data-driven approach to the problem, requiring only one photograph %
    and a user-specified target illumination as input (see Figure~\ref{fig:teaser}). 
    Our method relies on precomputed radiance transfer~\cite{sloan:2002:prt} (PRT) and spherical harmonics lighting~\cite{ramamoorthi2001efficient} (SH). Based on this, a convolutional neural network (CNN) decomposes the image into its albedo, illumination, and light transport components; from which the shading can be easily computed. Disentangling the illumination from all other factors in the scene allows for effective relighting, while the PRT-based scheme enables fast, efficient rendering.
    Our work lifts the assumption of Lambertian materials present in previous single-image human relighting methods~\cite{sengupta2018sfsnet, kanamori2018relighting}. We model the PRT decomposition in our framework by approximating material reflectance using an Oren-Nayar~\cite{oren1994generalization} and GGX microfacet model~\cite{walter:2007:microfacet} for the diffuse and specular components, respectively. In addition, we extend the image reconstruction formulation by adding a \emph{residual} term learned by our model, which 
    accounts for errors in image reconstruction that would be obtained using only the terms proposed by PRT.
    
    To train our model, we create a synthetic dataset containing almost 140,000 images with a rich variety of humans (105), poses (5), and illumination maps (266). We quantitatively and qualitatively evaluate relighting results on both synthetic images and real photographs, and perform extensive ablation experiments to validate our design choices in the model architecture, reflectance model for data generation, and loss functions.
    Compared with the current state of the art in full-body single-image human relighting~\cite{kanamori2018relighting}, our model yields more accurate reconstructions of relighted images for both synthetic images and real photographs. 
    We will make our code publicly available.

\section{Related Work}

\paragraph*{Inverse rendering} %
Single-image physically-based relighting typically requires solving an \textit{inverse rendering} problem where shape, reflectance, and illumination need to be inferred from a single image. This is a highly ill-posed problem, with infinite solutions, classically solved assuming that some information is known beforehand. \textit{Shape from shading}~\cite{ramachandran1988perception, ikeuchi1981numerical} is one of the earliest methods, estimating shape from shading under a known illumination. Other methods estimate shape relying on simple illumination models such as directional, point, or area light sources~\cite{christou1997light, okatani1997shape, lopez2009light}, or environmental lighting encoded into spherical harmonics~\cite{johnson2011shape}. Reflectance and illumination can be estimated from a known convex shape~\cite{chandraker2011image}, a shape with occluding contours~\cite{lopez2013multiple}, or just an approximated geometry~\cite{kholgade20143d}. A similar line of research has focused on \textit{intrinsic images}~\cite{barron2012color, ye2014intrinsic, garces2017intrinsic, garces2012intrinsic, weiss2001deriving}, which aims to decompose a scene into its shading and albedo components~\cite{land1971lightness}. 
Our method draws inspiration from intrinsic images, and we estimate albedo and shading from a single input image; however, we additionally  decompose shading into shape and illumination by developing a framework inspired by precomputed radiance transfer (PRT)~\cite{ramamoorthi2009precomputation, lehtinen2007framework, sloan:2002:prt}. In addition, our decomposition also takes into account diffuse and specular material reflections, thus producing more realistic results.

\paragraph*{Image-based rendering} A classic application of image-based rendering (IBR~\cite{shum2000review}) allows to take several pictures of a subject from the same viewpoint under different illuminations, and relight it using a weighted linear combination of those images~\cite{debevec2000acquiring, debevec1998efficient}. More sophisticated approaches optimize energy functions ~\cite{lempitsky2007seamless}, work with layered decompositions~\cite{sinha2012image}, or employ RGB-D cameras ~\cite{hedman2016scalable}. However, those techniques require a large number of input images, as well as precise control over the lighting, making them unfeasible for single-image, in-the-wild applications. 
Recent work exploits the potential of implicit representations and Fourier mappings of the input to learn high-quality 3D scene representations using one multi-layer perceptron (MLP) per scene and several hundreds of images as the input~\cite{mildenhall2020nerf,zhang2020nerf,boss2020nerd}, although these methods do not generalize across scenes. The work of Wang et al.~\cite{wang2021ibrnet} addresses this by combining implicit models with IBR to generate novel views without relighting. In contrast, our work develops a general framework that takes a single RGB image of a human as input, and outputs an intermediate representation suitable for relighting.

\paragraph*{Data-driven methods} 
Recent techniques leverage deep learning to predict illumination~\cite{hold2017deep, hold2019deep, gardner2017learning, lagunas2021joint}, estimate specular reflectance and illumination~\cite{georgoulis2017reflectance,  lombardi2015reflectance}, devise material reflectance metrics~\cite{delanoy2020role, lagunas2019similarity}, 
or perform intrinsic image decomposition~\cite{ma2018single, li2018learning,barron2014shape}.
For the particular case of humans, many relighting approaches rely on complex hardware setups~\cite{collet2015high,guo2019relightables, zhang2021neural}, which are not widely available; we instead focus on single RGB images as input.

Single-image human relighting approaches have been proposed for faces~\cite{wang2020single}: Sengupta et al.~\cite{sengupta2018sfsnet} show how we can relight faces using convolutional neural networks and spherical harmonics, later extended with more complex model architectures~\cite{zhou2019deep}, or by directly fitting encoder-decoder architectures to light-stage portrait data~\cite{sun2019single, nestmeyer2020learning}.
Closer to ours is the work of Kanamori and Endo~\cite{kanamori2018relighting}, performing full-body relighting. They use an encoder-decoder architecture to perform a single-image decomposition of the scene that is suitable for full-body human relighting. Our work lifts their assumption of materials being Lambertian by explicitly modeling the diffuse and specular reflectance in our data. We also add a residual term to the image reconstruction equation that allows to better model errors in the PRT image reconstruction.

\section{Background}
\label{sec:background}

In this section, we briefly review the building blocks of our technique: Spherical harmonics (SH) lighting~\cite{ramamoorthi2001efficient}, and precomputed radiance transfer (PRT)~\cite{sloan:2002:prt}.
    
PRT~\cite{sloan:2002:prt} and SH lighting~\cite{ramamoorthi2001efficient} enable rendering dynamic low-frequency environments with
realistic highlights and real-time shading. They estimate the amount of radiance reflected at a point in the scene by solving a simplified version of the rendering equation:
\begin{equation}
    R(x) = \int _{\mathbb S^2} L(x, \omega_i) \; T(x, \omega_i) \; \dwi,
    \label{eq:render}
\end{equation}
where $R$ is the reflected radiance or image intensity computed over the sphere $\mathbb S^2$ of incoming directions $\omega_i$, $L$ is the incoming light at point $x$ from direction $\wi$, and $T$ is a transport function computed for each vertex that includes the material reflectance $f_r$, visibility term $V$ that is 1 if the point is not occluded and 0 otherwise, and the cosine term which uses the normal $\textbf{n}$ at point $x$. The function $T$ can be expressed as:
\begin{equation}
    T(x, \omega_i) = f_r(x, \omega_i) \; V(x, \omega_i) \; (\omega_i \cdot \textbf{n}).
    \label{eq:transport}
\end{equation}

The formulation presented by PRT expands the illumination $L$ and the transport $T$ using (real) spherical harmonics basis functions $Y_{l,m}$, such that: 
\begin{equation}
\begin{split}
        L(x, \omega_i) &= \sum_{l=0}^\infty \sum_{m =-l}^l L_{l,m}(x) Y_{l,m}(\omega_i), \\  
        T(x, \omega_i) &= \sum_{l=0}^\infty \sum_{m =-l}^l T_{l,m}(x) Y_{l,m}(\omega_i) ,
\end{split}
\label{eq:sh_expansion}
\end{equation}
where $L_{l,m}$ and $T_{l,m}$ are the corresponding coefficients for illumination and transport, respectively 
(see~\cite[Sections 3 and 4]{ramamoorthi2009precomputation} for additional details on how to obtain $T_{l,m}$ and $L_{l,m}$). 
The integral in Equation~\ref{eq:render} then becomes:
\begin{equation}
    R(x)= \sum_{l=0}^\infty \; \sum_{m =-l}^l L_{l,m} (x) \; T_{l,m} (x).
    \label{eq:render_dot}
\end{equation}

This formulation has two advantages: It allows to approximate the rendering equation as a fast dot product, and it disentangles the illumination and the transport in the scene. In this way, relighting a scene only requires computing the coefficients of the new illumination $L'_{l,m}$, while keeping $T_{l,m}$ fixed.

Traditionally, relighting methods based on the estimation of illumination and transport coefficients from a single image soften the problem by assuming that the scene has a light source at a sufficient distance to neglect the angular variation between points, i.e., $L(x, \omega_i) \approx L(\omega_i)$. They also estimate a transport function $T$ encoding only the cosine term~\cite{sengupta2018sfsnet}, or the cosine term together with the visibility function~\cite{kanamori2018relighting}. These methods assume all materials to be Lambertian, removing the reflectance term from the transport $T_{l,m}(x)$, and modeling it as a constant for each point of the scene represented by the albedo~$\rho(x)$. 
With this, expressing $L_{l,m}$ as a vector $\mathbf L$ and $T_{l,m}(x)$ as a vector per point of the scene $\mathbf T(x)$, $R(x)$ can be approximated as (hereafter, we omit the dependency on $x$ for clarity):
\begin{equation}
    R \approx \underbrace{\rho}_{albedo} \cdot \; \underbrace{(\mathbf T^T \cdot \mathbf L)}_{shading \; S},
    \label{eq:render_prt}
\end{equation}
where the dot product between transport and illumination yields the shading $S(x)$ of a point in the scene, then scaled by the albedo $\rho$. The error of the approximation will be related to the number of coefficients used to estimate the illumination and transport in Equations~\ref{eq:sh_expansion} and~\ref{eq:render_dot}; this number of coefficients depends on the number of terms used to approximate the infinite term summation of Equation~\ref{eq:render_dot}, $l=[0..N]$. 

\begin{figure}
    \centering
    \includegraphics[width=0.95\linewidth]{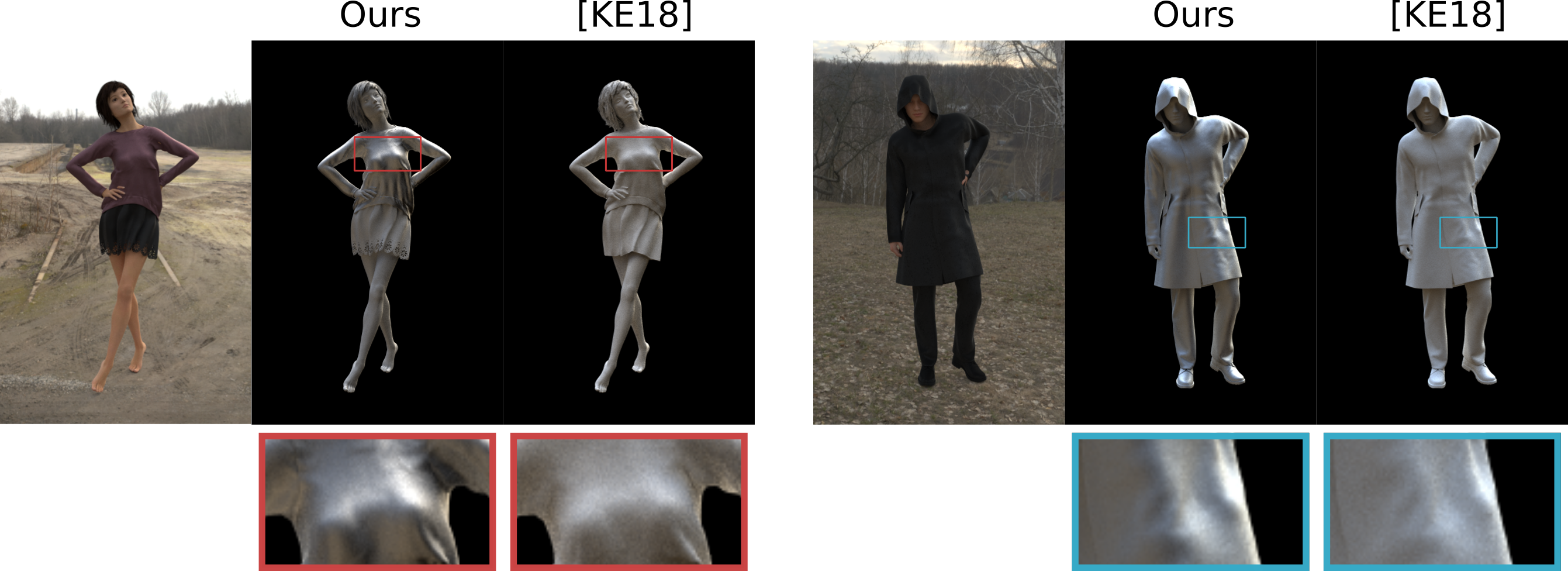}
    \caption{Comparison between the data generated with our framework and that of the recent work by Kanamori and Endo~\cite{kanamori2018relighting}, used to train the respective models. Our transport function $\mathbf T$ takes into account angular dependencies in the reflectance term, better capturing specular reflections and improving high-frequency details in the shading.}
    \label{fig:tport_render_cmp}
\end{figure}

To increase realism in the inferred and rendered images, we lift the Lambertian material assumption of previous work and include a better approximation of material reflectance in the transport function $\mathbf T$. We approximate the reflectance term in Equation~\ref{eq:transport} by keeping the albedo $\rho$ as a constant and using a white material with an Oren-Nayar~\cite{oren1994generalization} for the diffuse component, and a GGX model with Smith shadowing factor and Fresnel~\cite{walter:2007:microfacet} for the specular reflection. Then, we use Equation~\ref{eq:sh_expansion} to encode such reflectance in a new transport function $\mathbf T$, later used to render new images with Equation~\ref{eq:render_prt}.  As Figure~\ref{fig:tport_render_cmp} shows, this allows to better capture the directionality of specular reflections. Our reflectance model employs the following parameters: albedo, roughness, metallic, and transparency (refer to Section~\ref{sec:dataset} for additional details). Both the Oren-Nayar and the GGX models share the same roughness parameter. The final reflectance model is defined as a combination of up to seven BSDFs, which can be either a diffuse Oren-Nayar microfacets model or a specular GGX model.
\section{Our Image Reconstruction Formulation}
\label{sec:framework}

This section describes our image reconstruction formulation, including the motivation behind the addition of a new residual term.

Since using a large number of basis coefficients in Equation~\ref{eq:render_prt} to approximate $R$ with a low error is computationally expensive, we introduce an additional residual vector $\textbf{E}$, leading to:
\begin{equation}
    R \approx \underbrace{\rho}_{albedo} \cdot \; \underbrace{(\mathbf T^T \cdot \mathbf L)}_{shading \; S} + \underbrace{(\mathbf E^T \cdot \mathbf L)}_{residual \; E},
    \label{eq:render_prt_residual}
\end{equation}
where the dot product between the residual vector $\textbf{E}$ and the illumination $\textbf{L}$ yields a residual value per point $E(x)$. Again, the dependency on $x$ is omitted for clarity, but Equation~\ref{eq:render_prt_residual} applies to each point in the scene, yielding the corresponding images; in the following, we will use $\textrm{S}$ to denote the shading image, and $\textrm{E}$ for the residual image.
The residual vector $\textbf{E}$ does not have a physical meaning; instead, it is a set of learned coefficients that aim to model the errors in image reconstruction that we would obtain using only the terms (albedo, transport, and illumination) with a limited number of coefficients.

\subsection{Problem Formulation}
\label{sec:problem_description}
Our main goal is to relight an image $\psi$ with a full-body human in it, given a user-specified target illumination $\mathbf L'$: 
\begin{equation}
     \hat\psi = \mathcal R(\psi, \mathbf {L}'),
\end{equation}
where $\mathcal R$ is a relighting function, and $\hat\psi$ is the resulting relighted image with target illumination $\mathbf L'$. 

Using a model such as the one in Equation~\ref{eq:render_prt_residual}, one can change $\mathbf{L}$ to $\mathbf{L}'$ to obtain the relighted image.
However, given a single image as input, the transport $\mathbf{T}$, illumination $\mathbf{L}$, residual $\textbf{E}$, and albedo $\rho$ are unknown. 
To obtain an approximation of $\mathbf{T}$, $\mathbf{L}$, $\textbf{E}$, and $\rho$, we introduce the parametric function $\mathcal G$, which takes as input the image $\psi$ and a set of parameters $\beta$:
\begin{equation}
    \{ \textbf{T}, \textbf{L}, \textbf{E}, \rho\} \approx \mathcal G(\psi, \beta).
\end{equation}
In particular, we model $\mathcal G$  using a convolutional neural network whose parameters are represented by $\beta$. Note that $\mathcal G$ tries to approximate each of the terms  $\{ \textbf{T}, \textbf{L}, \textbf{E}, \rho\}$ irrespective of the underlying reflectance model previously used to generate them. With the output of $\mathcal G$ and a given user-specified illumination $\mathbf {L}'$, we can use Equation~\ref{eq:render_prt_residual} to directly approximate the relighting function $\mathcal R$.

\section{Dataset}
\label{sec:dataset}
To learn the parametric function $\mathcal G$ introduced in Section~\ref{sec:framework} we have created a synthetic human image dataset of almost 140,000 images including a rich variety of humans, poses, and illuminations, which we describe in this section.

\paragraph*{Human 3D models} Existing models captured using photogrammetry mostly consist only of diffuse and normal maps. To fully exploit the capabilities of our framework and go beyond Lambertian materials, we purchase rigged 3D human models and clothing from the \textit{DAZ} website~\cite{daz3d}, which include realistic materials and texture maps for diffuse color, specular, opacity, roughness, metallic, translucency, and normals.
In total, we collected 105 different clothed models; augmented with five poses each, this yields a total of 525 different renditions.
For each pose we simulate cloth interaction after posing the model, and, to foster diversity, perform subtle random changes to the hue of the diffuse color.

\paragraph*{Illumination maps} 
We used freely-available spherical high-dynamic range images (HDRIs) from HDRIHaven~\cite{hdrihaven}, corresponding to both indoor and outdoor scenarios. To normalize the HDRIs, we compute a \textit{reference radiance} for each image by obtaining the mean shading in Equation~\ref{eq:render_prt}, where $\textbf{L}$ are the coefficients of the HDRI, and $\textbf{T}$ is obtained analytically by sampling all unit directions in the sphere. We scale all the illuminations $\textbf{L}$ to have a reference radiance in the range $[0.7, 0.9]$. In total we gathered 266 different HDRIs.

\begin{figure}[t!]
    \centering
    \includegraphics[width=0.95\linewidth]{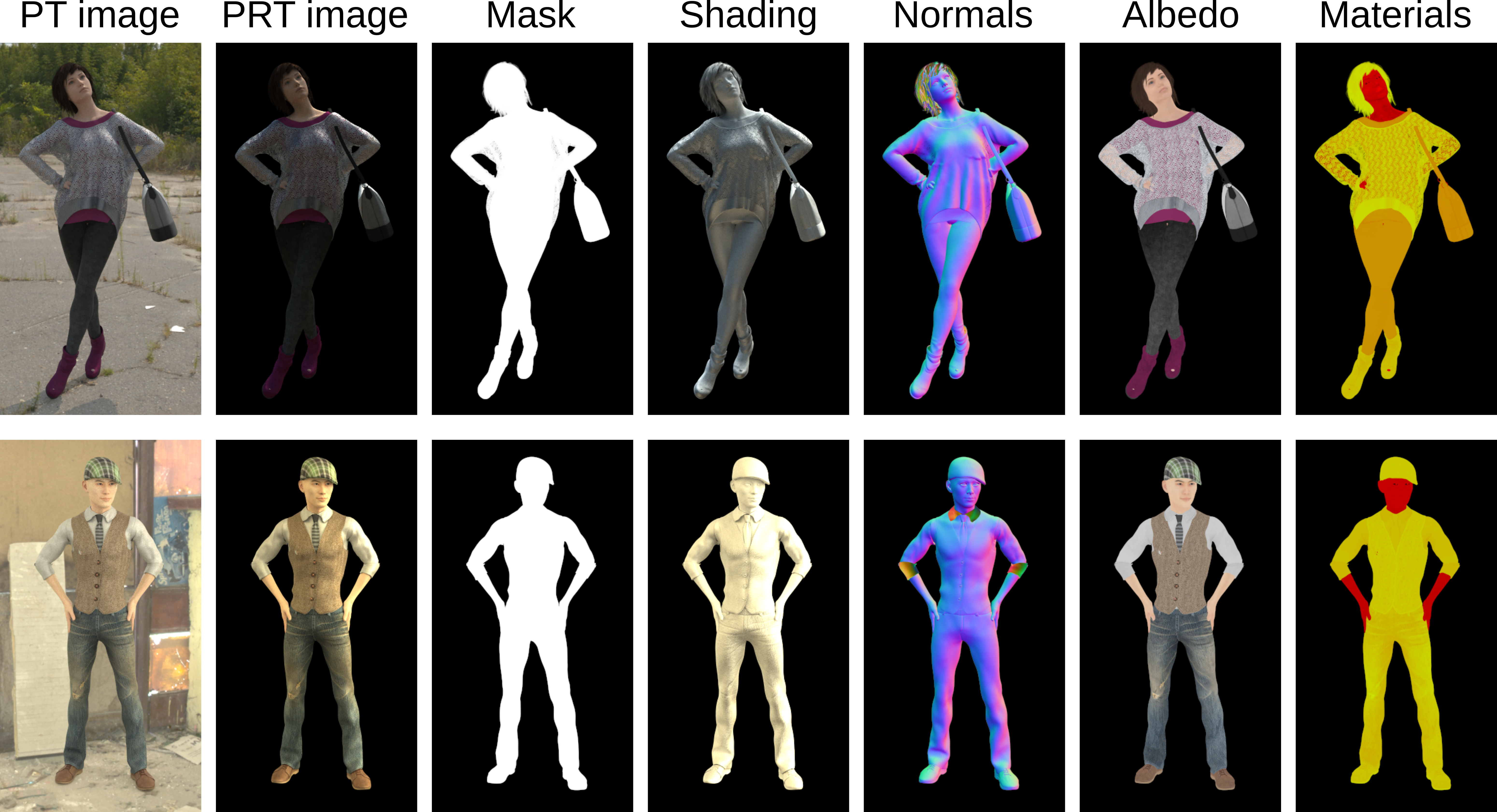}
    \caption{
    Two examples in our dataset. For each scene we obtain its path-traced (PT) rendered image, its PRT image rendered with our image reconstruction formulation, the alpha mask, the shading, the normals, the albedo, and a material map describing the roughness, transparency, and metallic (each encoded in a separate channel of an RGB image).}
    \label{fig:dataset-teaser}
\end{figure}

\paragraph*{Rendering} We used Monte Carlo path tracing to render realistic images and to obtain the transport vector $\mathbf T$ for each scene. To generate $\mathbf L$ for each illumination, we integrate over the unit sphere of directions. We fix $N=4$ ($l=[0..4]$), which leads to 25 spherical harmonics coefficients in $\mathbf T$ and $\mathbf L$ (in contrast, the work of Kanamori and Endo~\cite{kanamori2018relighting} estimates only Lambertian materials and uses $N=2$). Among all the available maps defining reflectance for each purchased model, during rendering we employ the albedo (diffuse color), roughness, metallic, and transparency maps. In total, we render 139,650 different scenes. For each scene, we generate: Its path-traced (PT) image, the PRT image computed using Equation~\ref{eq:render_prt},
an alpha mask of the human, the shading, the normals, the albedo, and a material map containing the roughness, transparency, and metallic, each of them encoded in separate channel of an RGB image. All images are rendered with a resolution of $768 \times 768$ pixels; using 256 samples per pixel for the PT image, and 1,024 for the transport $\mathbf T$ and all other scene properties. Figure~\ref{fig:dataset-teaser} shows two samples from our dataset, cropped down from the squared aspect ratio.

\section{Our Model}
\label{sec:architecture}
 In this section we explain our model architecture and its components, together with an intuition behind our design choices; in addition, we provide details on our training, hyper-parameters, and loss function.

\subsection{Model Architecture}
\begin{figure}[t!]
        \centering
        \includegraphics[width=0.95\linewidth]{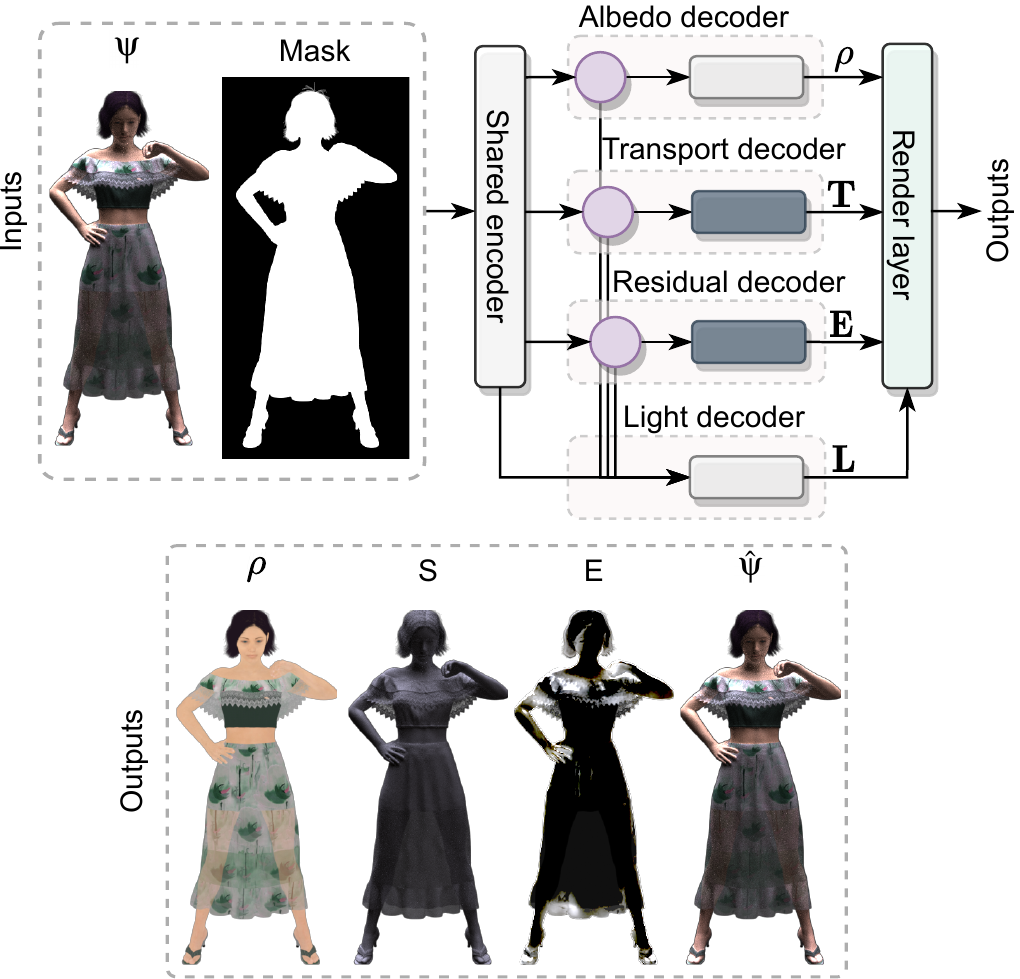}
        \caption{Our model architecture. The masked input image goes through a shared encoder that converts it into a feature map. Such feature map simultaneously serves as an input to the albedo, transport, and residual decoders. The three decoders output the albedo $\rho$, transport $\textbf{T}$, and residual coefficients $\textbf{E}$, respectively. The features from these three decoders and from the shared encoder are concatenated and fed to the light decoder, which outputs the illumination coefficients $\textbf{L}$. Last, the rendering layer outputs the albedo (equal to the output from the albedo decoder), shading, residual image, and the final relighted image.}
        \label{fig:framework-workflow}
    \end{figure}

     To represent our parametric function $\mathcal G$ we use a convolutional neural network based on a UNet-like model~\cite{ronneberger2015u}. Figure~\ref{fig:framework-workflow} shows an overview. It consists of a shared encoder that receives the input image $\psi$, and several decoders responsible for estimating albedo $\rho$, transport $\textbf{T}$, residual coefficients $\textbf{E}$, and the illumination of the input image $\textbf{L}$. We add skip-connections between the shared encoder and each decoder to encourage better reconstructions, except for the light decoder. Last, we have a rendering layer based on Equation~\ref{eq:render_prt_residual} that generates the shading $\text{S}$, the residual $\text{E}$, and the final relighted image $\hat\psi$.
    
    \paragraph*{Shared encoder} Our encoder has a standard architecture consisting of several convolutional blocks with batch-normalization (BN) that sequentially decrease the resolution of the features by a factor of two. The features between convolutional blocks are used as skip-connections with the decoders. 
    
\begin{figure}
    \centering
    \includegraphics[width=0.95\linewidth]{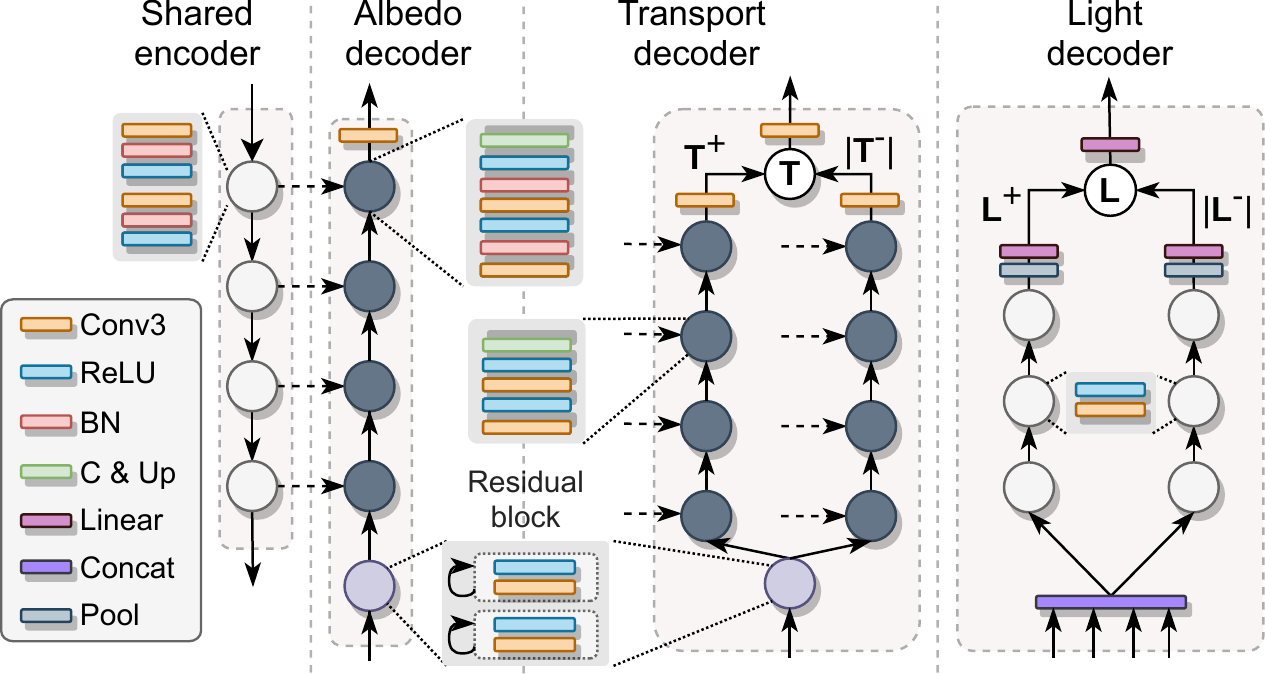}
    \caption{Workflow of each component of our model. The shared encoder contains several convolutional blocks that reduce the spatial dimensions by two and output a feature map of the input. Such feature map goes to the albedo, transport, and residual decoders. Each decoder (except for the light one) has a residual block and a generator. The generator concatenates skip-connections and upscales (C \& Up) the spatial resolution of the features. The output of the decoders has the same spatial resolution as the input image. Last, the light decoder uses the features of the encoder, together with the features from the residual block of each decoder, to predict the illumination in the scene. 
    }
    \label{fig:encoder-workflow}
    \end{figure}

    \paragraph*{Decoders} Each decoder has a \textit{residual block} (similar to ResNet~\cite{he2016deep}), and a \textit{generator block} except for the light decoder that only has a generator block. The generator block varies between decoders. The output of the albedo, transport, and residual coefficients decoder has the same spatial resolution as the input image. We only add batch-normalization to the albedo decoder. The architecture of each generator is as follows (see also Figure~\ref{fig:encoder-workflow}): 
    
    \begin{itemize}
        \item The albedo decoder has several convolutional blocks with skip-connections. In each convolutional block features are scaled by a factor of two. The output of the albedo decoder is clamped to lie in the range $[0, 1]$. 
        
        \item To properly capture geometry and material reflectance in the scene, a good estimation of the transport matrix $\textbf{T}$ is needed. The transport and residual decoders feature a generator tailored for the PRT decomposition in Equation \ref{eq:render_prt_residual}. Deep neural networks, by design, add non-linear functions that clamp negative values. However, the transport coefficients are defined with both positive and negative values. Thus, we would rely on the last convolution without non-linearities to generate all the negative content in the coefficients. To give additional degrees of freedom to the decoders, we decompose the coefficients as $\textbf{T} = \textbf{T}^+ - |\textbf{T}^-|$ where $\textbf{T}^+$ corresponds to the positive part and $|\textbf{T}^-|$ is the absolute value of the negative part. Instead of directly predicting $\textbf{T}$, we add two generators (similar to the albedo one) to predict $\textbf{T}^+$ and $|\textbf{T}^-|$, respectively, and later we reconstruct the coefficients $\textbf{T}$. We apply a similar strategy to the residual coefficients $\textbf{E}$. 
        \item The light decoder differs from the previous as its input is the output of the shared encoder and the residual blocks of the albedo, transport, and residual decoders. Those features go straight to a generator that follows a similar decomposition as for the transport and residual decoder, however, the generator architectures differ. The generator has several convolutional blocks that reduce the spatial dimensions of the features by a factor of two. After the convolutions, we perform an average pooling making the features one-dimensional, and a fully-connected layer outputs the positive and negative illumination coefficients in each generator, with shape $3*25$ (25 being the total number of coefficients when $N=4$). Then, we reconstruct $\textbf{L}$ using the positive and negative part.
    \end{itemize}

\subsection{Training}
\label{subsec:training}
    The dataset in Section~\ref{sec:dataset} is split into training and validation, where we select 7 clothed models (with all their poses) that are representative of challenging scenes as the validation set. The rest of the humans with their poses are used for training.
    The input to our model are images rendered with PRT, where we crop the human using the bounding-box defined by the mask with a padding of 20 pixels.  Since our network is fully-convolutional it allows inputs of arbitrary resolution. We normalize the image pixels to lie in the range $[-1, +1]$ and multiply it by the alpha mask before forwarding it through the model.  For training we use the Adam optimization algorithm~\cite{kingma2014adam} with the decoupled weight decay regularization~\cite{loshchilov2017decoupled}. The learning rate has a value of $5 \cdot 10^{-5}$. We set an effective batch size of 16. We use the PyTorch framework~\cite{pytorch2019neurips} with PyTorch-Lightning~\cite{falcon2019pytorch} to design our model and experiments. The model is trained for 25 epochs on 
    eight Tesla V100-SXM2-16GB, lasting 55 hours approximately.

    \begin{figure*}[!t]
        \centering
        \includegraphics[width=0.9\linewidth]{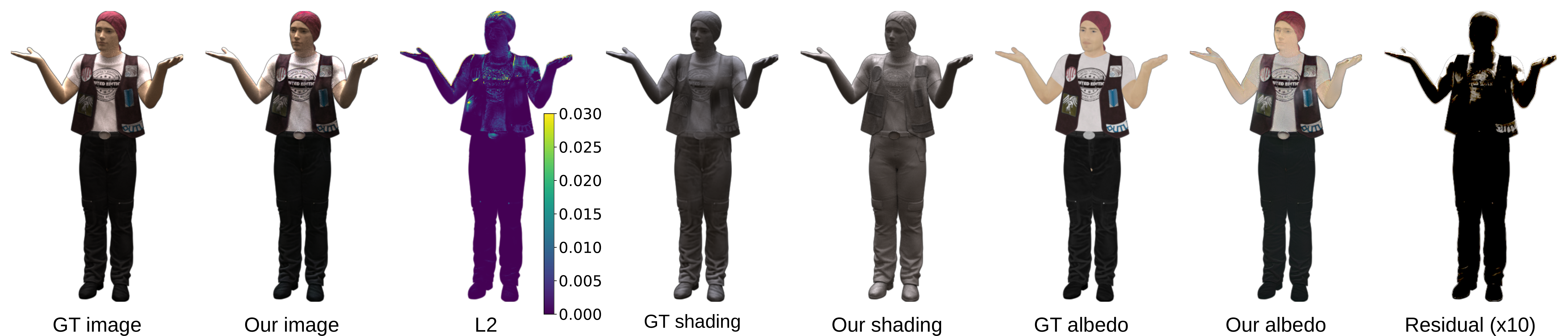}
        \caption{Example result of our model for a synthetic image (see also Table~\ref{tab:results_synthetic}, synthetic images). Neither the human nor the illumination were used for training. 
        We show direct comparisons with the ground truth (GT), the L2 error in the final image, and our residual term scaled by a factor of 10 for visualization purposes. 
        }
        \label{fig:results_synthetic}
    \end{figure*}
    \begin{figure}[!t]
        \centering
        \includegraphics[width=\linewidth]{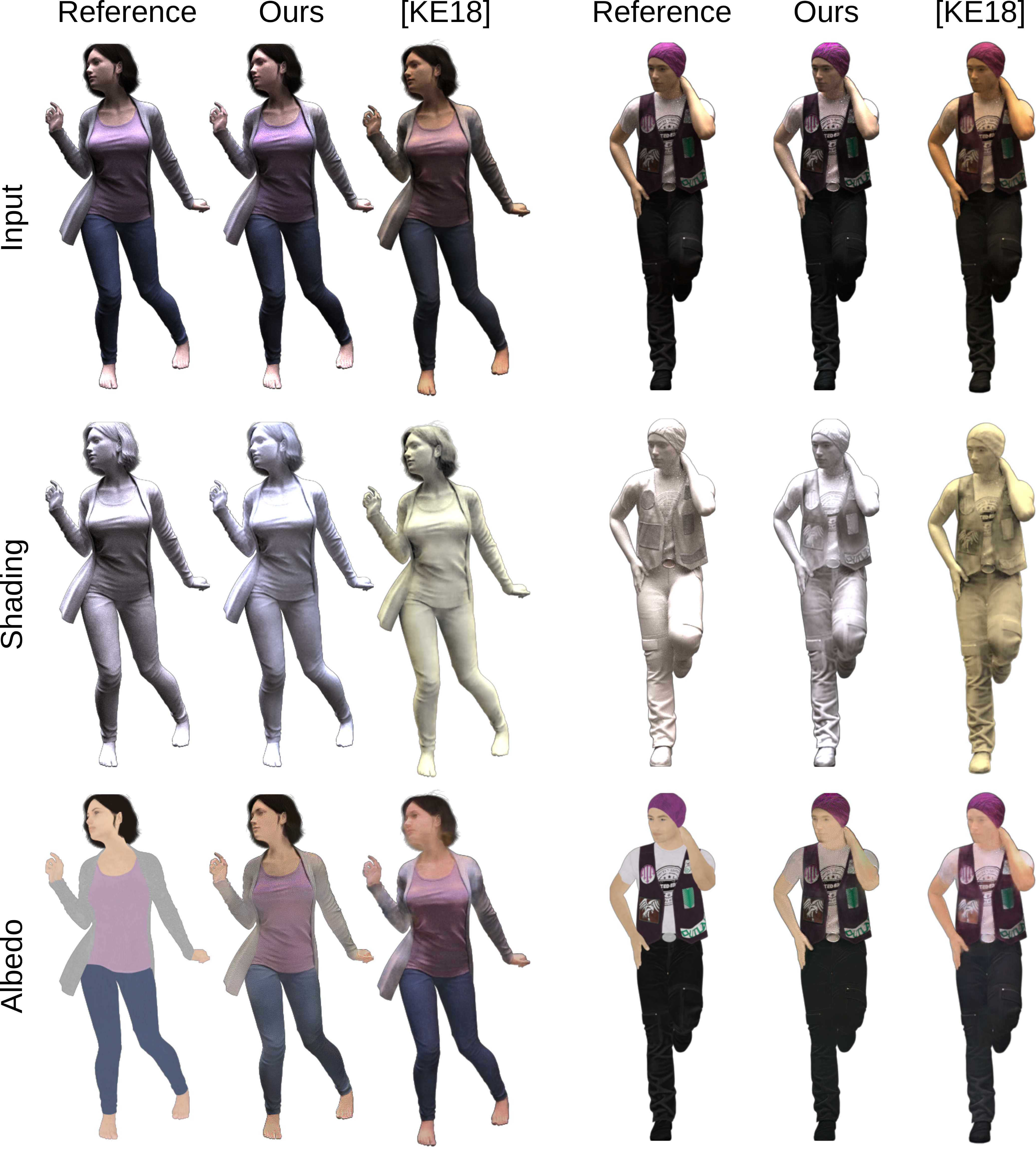}
        \caption{Comparison between our model and the model provided by Kanamori and Endo~\cite{kanamori2018relighting} in two examples of the validation dataset.
        We can see how our model outperforms them rendering the input image, albedo, and shading. Note that the shading encodes both the transport and the illumination of the scene.
        }
        \label{fig:results_synthetic_comp}
    \end{figure}
    \begin{figure*}[!ht]
        \centering
        \includegraphics[width=\linewidth]{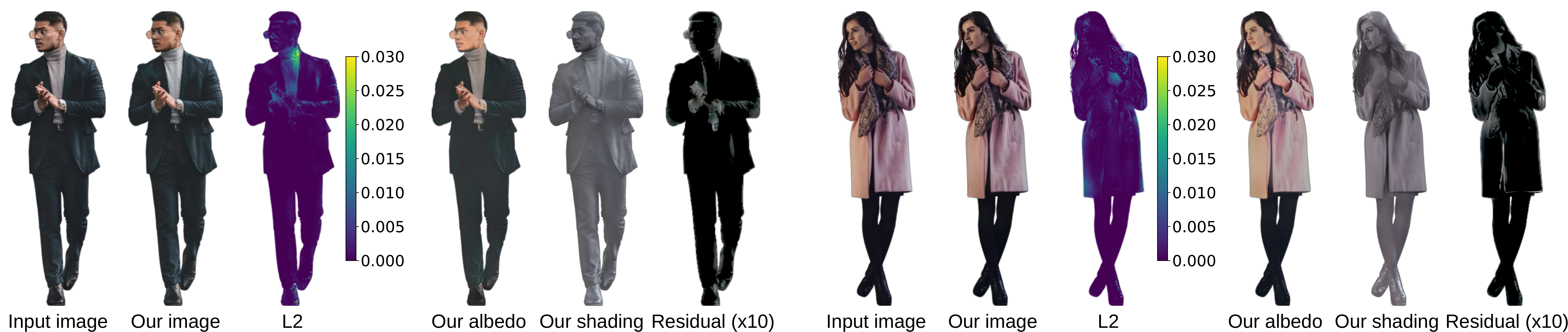}
        \caption{Example results of our model on real photographs (see also Table~\ref{tab:results_synthetic}, real photographs). 
        For each image, from left to right: Ground truth input image, resulting image relighted with our model, L2 error, albedo, shading, and residual term scaled by a factor of 10 for visualization purposes. 
        }
        \label{fig:results_real}
    \end{figure*}

\subsection{Loss Functions}
Our loss function $\mathcal L$ can be expressed as:
   \begin{equation}
        \mathcal L =\mathcal L_{\rho} + \mathcal L_{\textbf{T}} + \mathcal L_{\textbf{L}} +\mathcal L_{\text{S}} + \mathcal L_{\hat{\psi}}.
    \end{equation}
where each term supervises the prediction of albedo,  transport, illumination, shading, and the final relighted image. Note that the residual coefficients are not directly supervised. Instead, we let the network freely learn a set coefficients $\textbf{E}$ that aim to improve the quality of the rendered images. Each of the terms in $\mathcal L$ is additionally composed of different losses. We linearly combine the different terms using a weight of 1 for all of them.
    \begin{itemize}
        \item \textbf{Reconstruction loss ($\mathcal L_{L1}$)} 
        We apply an L1 loss function to each predicted map with respect to ground-truth data. To encourage a better reconstruction, we leverage the architecture tailored for PRT rendering, and additionally include an L1 loss between the positive and negative coefficients in $\mathcal L_{\textbf{T}}$ and $\mathcal L_{\textbf{L}}$.
        \item \textbf{Render loss ($\mathcal L_{r}$)} 
        The terms in Equation~\ref{eq:render_prt_residual} are computed using the albedo, transport, illumination, and residual vectors. 
        For each of those vectors (except the residual $\mathbf{E}$), there is both a predicted (which is being learned) and a ground truth vector.
        To increase robustness, we introduce in $\mathcal L_{\text{S}}$ and $\mathcal L_{\hat\psi}$ an L1 error term 
        for each possible way of generating the shading and relighted image in Equation~\ref{eq:render_prt_residual} from the predicted and ground truth vectors.
        \item \textbf{Log loss ($\mathcal L_{\log}$)} 
        The transport, and the illumination coefficients have an unbounded range. To compress it, we apply a logarithmic loss of the following form: $$\mathcal L_{\log}= ||\log(|x|+1) - \log(|\hat{x}| + 1)||_2^2$$ in $\mathcal L_{\textbf{T}}$ and $\mathcal L_{\textbf{L}}$. We apply $|x|$ in the logarithmic loss to avoid errors on the negative values of the coefficients. We leverage the PRT decomposition to apply the logarithmic loss also to the positive and negative decomposition of transport and illumination. 
    \end{itemize}

\section{Results}
\label{sec:results}
    
We show and evaluate results of our model on both synthetic images, where ground truth data is available, and real photographs. 
Throughout the evaluation, we show the reconstructed albedo $\rho$, shading $\text S$ (resulting from the combination of transport $\mathbf{T}$ and target illumination $\mathbf{L}'$, see Equation~\ref{eq:render_prt_residual}), the final rendered result $\hat\psi$, and the residual image $\text E$. We also include ablation studies to clearly demonstrate the influence of each component in the final relighted images. 
   \begin{table*}[!t]
    \center
    \caption{Quantitative results of our model for synthetic images and real photographs, measured with three metrics: L1 and L2 distances, and PSNR. Note that the L1 and L2 metrics have been scaled by a factor of 100. We also include a comparison to the model of Kanamori and Endo~\cite{kanamori2018relighting}, which our model consistently outperforms. Boldface highlights the best result in each case. }
    \label{tab:results_synthetic}
    \resizebox{\linewidth}{!}{\begin{tabular}{l|ccc|ccc|ccc|ccc}
    \toprule
          & \multicolumn{9}{c|}{\textsc{Synthetic images}}     & \multicolumn{3}{c}{\textsc{Real photographs}}      \\                                                                                                                                      
        \cmidrule(r{0.2em}){2-10}      \cmidrule(r{0.2em}){11-13}    
                                    & \multicolumn{3}{c}{\textsc{Albedo}}                   & \multicolumn{3}{c}{\textsc{Shading}}              & \multicolumn{3}{c|}{\textsc{Image}}  & \multicolumn{3}{c}{\textsc{Image}}                   \\ 
        \cmidrule(r{0.2em}){1-1}      \cmidrule(r{0.2em}){2-4}                                \cmidrule(r{0.2em}){5-7}                               \cmidrule(r{0.2em}){8-10} \cmidrule(r{0.2em}){11-13}
        Model                       & L1 (x100)         & L2 (x100)       & PSNR            & L1 (x100)        & L2 (x100)       & PSNR             & L1 (x100)        & L2 (x100)       & PSNR     & L1 (x100)                  & L2 (x100)              & PSNR        \\ 
        \cmidrule(r{0.2em}){1-1}      \cmidrule(r{0.2em}){2-4}                                \cmidrule(r{0.2em}){5-7}                               \cmidrule(r{0.2em}){8-10} \cmidrule(r{0.2em}){11-13}
        \textbf{Ours}                        & \textbf{2.88}     & \textbf{0.44} & \textbf{24.18}    & \textbf{3.77} & \textbf{0.71} & \textbf{24.05}    & \textbf{1.64} & \textbf{0.19} & \textbf{28.94}     & \textbf{1.17}      & \textbf{0.08}  & \textbf{31.42}     \\
        Kanamori and Endo           & 4.95              & 1.19          & 20.68             & 6.75          & 1.90          & 18.29             & 2.94          & 0.47          & 26.06         & 2.14       & 0.20   & 28.38              \\ 
        \bottomrule
    \end{tabular}}
\end{table*}
\begin{figure}[!ht]
    \centering
    \includegraphics[width=0.95\linewidth]{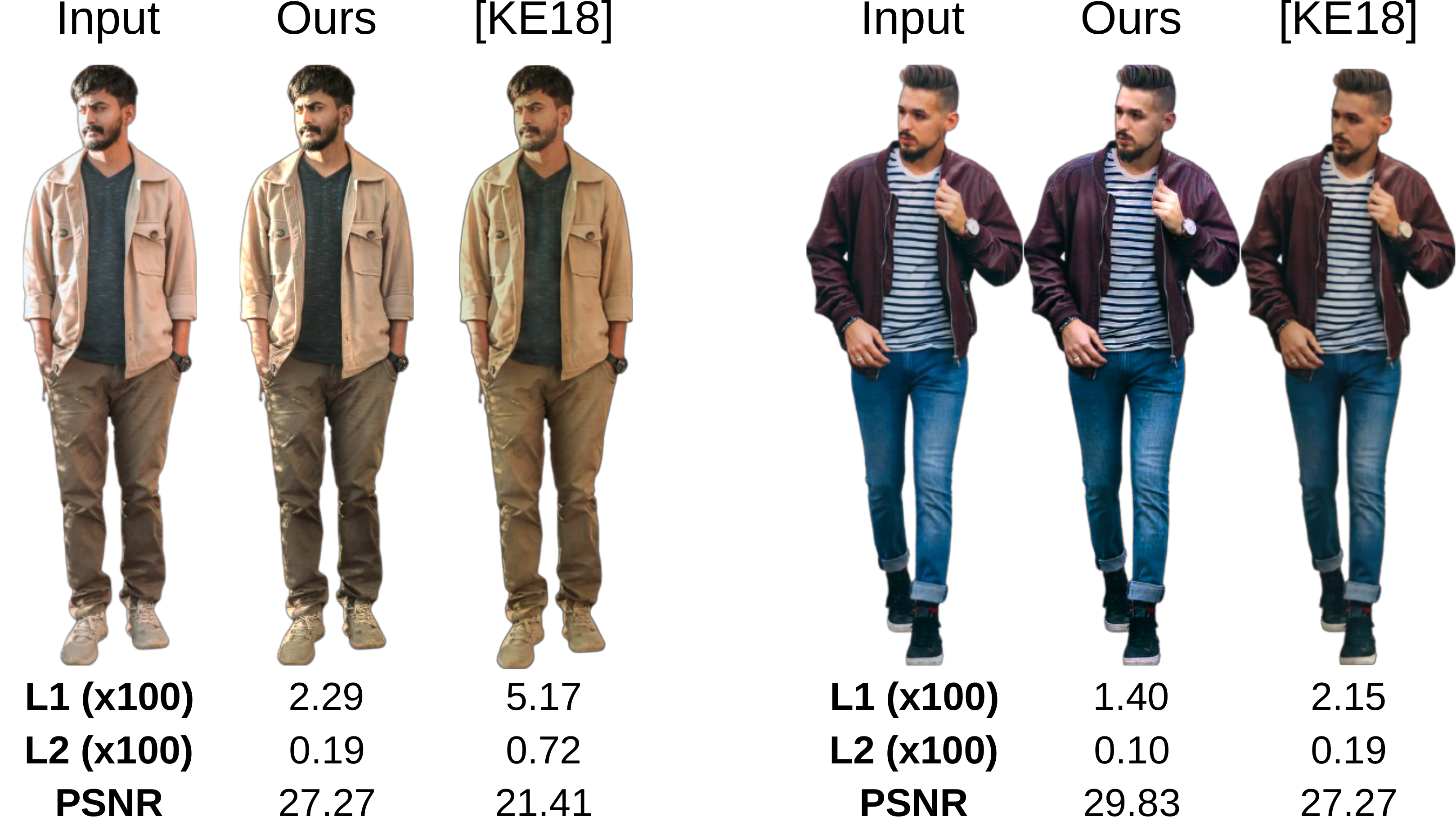}
    \caption{Image reconstructions obtained by our model, and the model provided by Kanamori and Endo~\cite{kanamori2018relighting}. We can see how our model outperforms them in the three metrics (see also Table~\ref{tab:results_synthetic}, real photographs). Note that the L1 and L2 metrics have been scaled by a factor of 100. In addition, our model better captures skin and cloth albedo, and the directionality of the illumination. 
    }
    \label{fig:outofdata_render_comp}
\end{figure}
\begin{figure}[!ht]
    \centering
    \includegraphics[width=\linewidth]{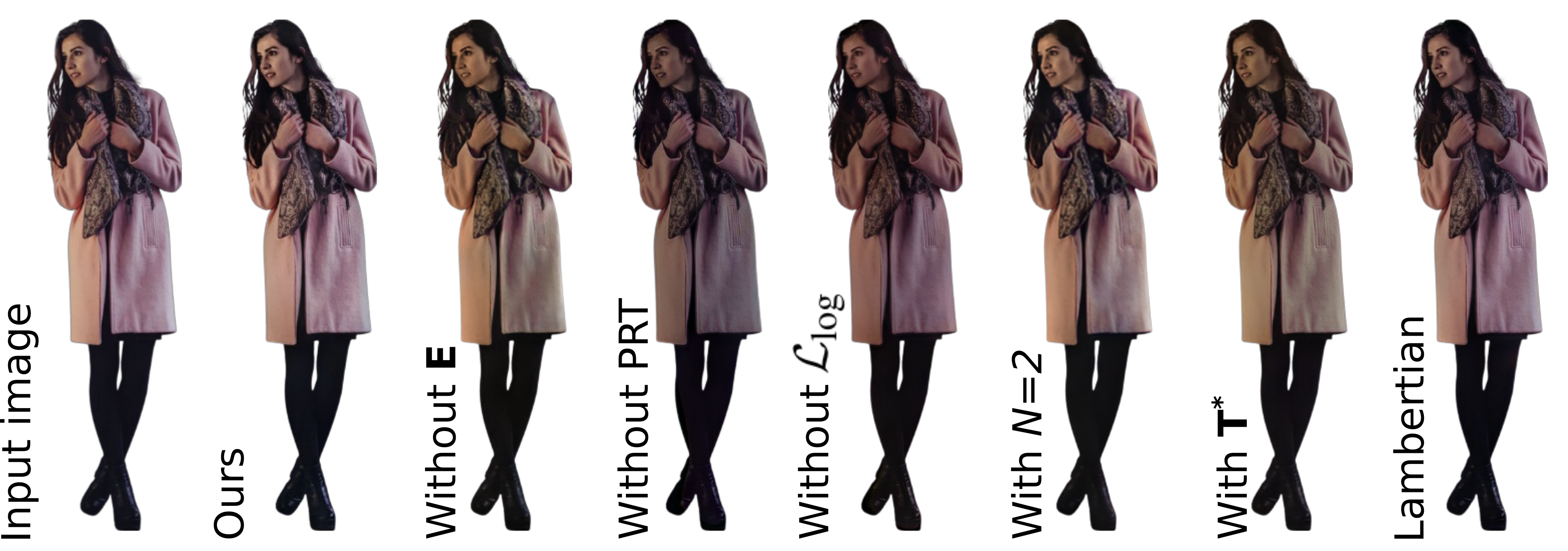}
    \caption{Reconstruction results obtained on the different ablation experiments. We can clearly observe how our full model better captures the appearance of the input photograph.}
    \label{fig:ablation_real}
\end{figure}

\paragraph*{Synthetic images} 
We use the validation subset of our dataset (see Section~\ref{subsec:training}) rendered with six new illuminations not used for training:  \emph{ennis}, \emph{grace}, \emph{pisa}, \emph{doge}, \emph{glacier} and \emph{uffizi}~\cite{devebecHDR}.  
We render the final relighted (target) image using the predicted illumination of the scene to reconstruct the shading and the residual. 
Since ground truth data is available, we also compute quantitative error measures for the albedo, shading, and final rendered image. Specifically, we compute the L1 and L2 distances, as well as PSNR, averaged across the dataset. Table~\ref{tab:results_synthetic} (synthetic images) shows the results, including a comparison with the pretrained model of the recent work by Kanamori and Endo~\cite{kanamori2018relighting}. 
Our more complete material reflectance formulation, together with our residual term (see ablation studies in Subsection~\ref{subsec:ablation}) lead to significantly lower L1 and L2 values, and higher PSNR for the albedo and shading, as well as the final relighted image.
Figure~\ref{fig:results_synthetic} shows a direct comparison of our reconstructed image with the ground truth; both images match with a very small L2 error.  Figure~\ref{fig:results_synthetic_comp} shows a comparison between our model and the pretrained model given in the work of Kanamori and Endo on synthetic images. We can see how our model better estimates the shading and albedo, leading to more accurate results where directional effects are better reproduced (see the highlights in the face of the first image, for instance).

\begin{table*}[!htbp]
    \center
    \caption{Quantitative results of our model and all the ablation experiments for synthetic images and real photographs. We measure three different metrics: L1 and L2 distances, and PSNR. Note that the L1 and L2 metrics have been scaled by a factor of 100. Our model outperforms all other experiments. Boldface highlights the best result in each case.}
    \label{tab:ablation_real_and_synthetic}
    \resizebox{\linewidth}{!}{\begin{tabular}{l|ccc|ccc|ccc|ccc}
        \toprule   
                                    &                                            \multicolumn{9}{c|}{\textsc{Synthetic images}}                                                                      & \multicolumn{3}{c}{\textsc{Real photographs}}    \\  
                 \cmidrule(r{0.2em}){2-10}                                                   \cmidrule(r{0.2em}){11-13}

                                    & \multicolumn{3}{c}{\textsc{Albedo}}                   & \multicolumn{3}{c}{\textsc{Shading}}              & \multicolumn{3}{c|}{\textsc{Image}}                & \multicolumn{3}{c}{\textsc{Image}}                                      \\ 
        \cmidrule(r{0.2em}){1-1}      \cmidrule(r{0.2em}){2-4}                                \cmidrule(r{0.2em}){5-7}                               \cmidrule(r{0.2em}){8-10}                       \cmidrule(r{0.2em}){11-13}
        Model                       & L1 (x100)           & L2 (x100)        & PSNR            & L1 (x100)         & L2 (x100)        & PSNR             & L1 (x100)         & L2 (x100)        & PSNR      & L1  (x100)        & L2 (x100)        & PSNR\\ 
        \cmidrule(r{0.2em}){1-1}      \cmidrule(r{0.2em}){2-4}                                \cmidrule(r{0.2em}){5-7}                               \cmidrule(r{0.2em}){8-10} \cmidrule(r{0.2em}){11-13}
        \textbf{Ours}                                      & \textbf{2.88}     & \textbf{0.44} & \textbf{24.18}    & \textbf{3.77} & \textbf{0.71} & \textbf{24.05}    & \textbf{1.64} & \textbf{0.19} & \textbf{28.94}  & \textbf{1.17} & \textbf{0.08}  & \textbf{31.42} \\
        Without $\textbf{E}$                      & 3.67              & 0.66          & 23.05             & 6.71           & 2.74          & 17.89            & 1.97          & 0.24          & 27.13           & 2.64          & 0.29           & 26.24      \\
        Without PRT decomposition                 & 4.54              & 1.00          & 21.08             & 10.57          & 5.69          & 14.43            & 2.13          & 0.24          & 26.80           & 2.55          & 0.30           & 26.66     \\
        Without $\mathcal L_{\log}$               & 4.02              & 0.83          & 21.84             & 10.34          & 5.35          & 14.55            & 2.14          & 0.24          & 27.82           & 2.31          & 0.25           & 26.85  \\ 
        With $N=2$                                & 3.31              & 0.58          & 23.33             & 8.60           & 4.21          & 16.18            & 1.83          & 0.22          & 28.33           & 2.08          & 0.18           & 27.76  \\   
        With $\textbf{T}^*$                    & 3.68        & 0.76    & 21.74       & 7.53     & 3.54    & 16.65      & 2.25    & 0.31    & 27.29     & 1.75    & 0.14     & 29.43  \\   
        Lambertian materials                          & 3.58    & 0.68    & 22.66       & 7.22     & 2.91    & 17.09      & 1.91    & 0.21    & 27.92     & 1.82    & 0.18     & 29.15  \\   
\bottomrule
    \end{tabular}}
\end{table*}

\paragraph*{Real photographs} 
To test our model on real photographs we use free-license images downloaded from Unsplash~\cite{unsplash}. To obtain the alpha mask we rely on freely available APIs~\cite{removebg}. In total we collected 10 different images with a single human in them. 
Error metrics for the resulting rendered images, averaged over the 10 photos, can be found in Table~\ref{tab:results_synthetic} (real photographs). As with synthetic images, our results significantly outperform previous work~\cite{kanamori2018relighting}.
Maybe surprisingly, the error metrics indicate better results with real photographs (both for our method and using the pretrained model of Kanamori and Endo) than using synthetic images. This is possibly due to the fact that the synthetic validation dataset contains some quite extreme illuminations (e.g., \emph{glacier} or \emph{grace}), while the photographic dataset has more natural illuminations that the two models are able to reproduce better. Figure~\ref{fig:results_real} shows the reconstruction performed by our model for two different input photographs, including albedo and shading components, while a direct  comparison with previous work is shown in Figure~\ref{fig:outofdata_render_comp}. Again, we see how our model is able to better capture directional effects (see, e.g., the faces or the highlights in the jackets) and overall produce more accurate reconstructions.

\begin{figure*}[!t]
    \centering
    \includegraphics[width=\linewidth]{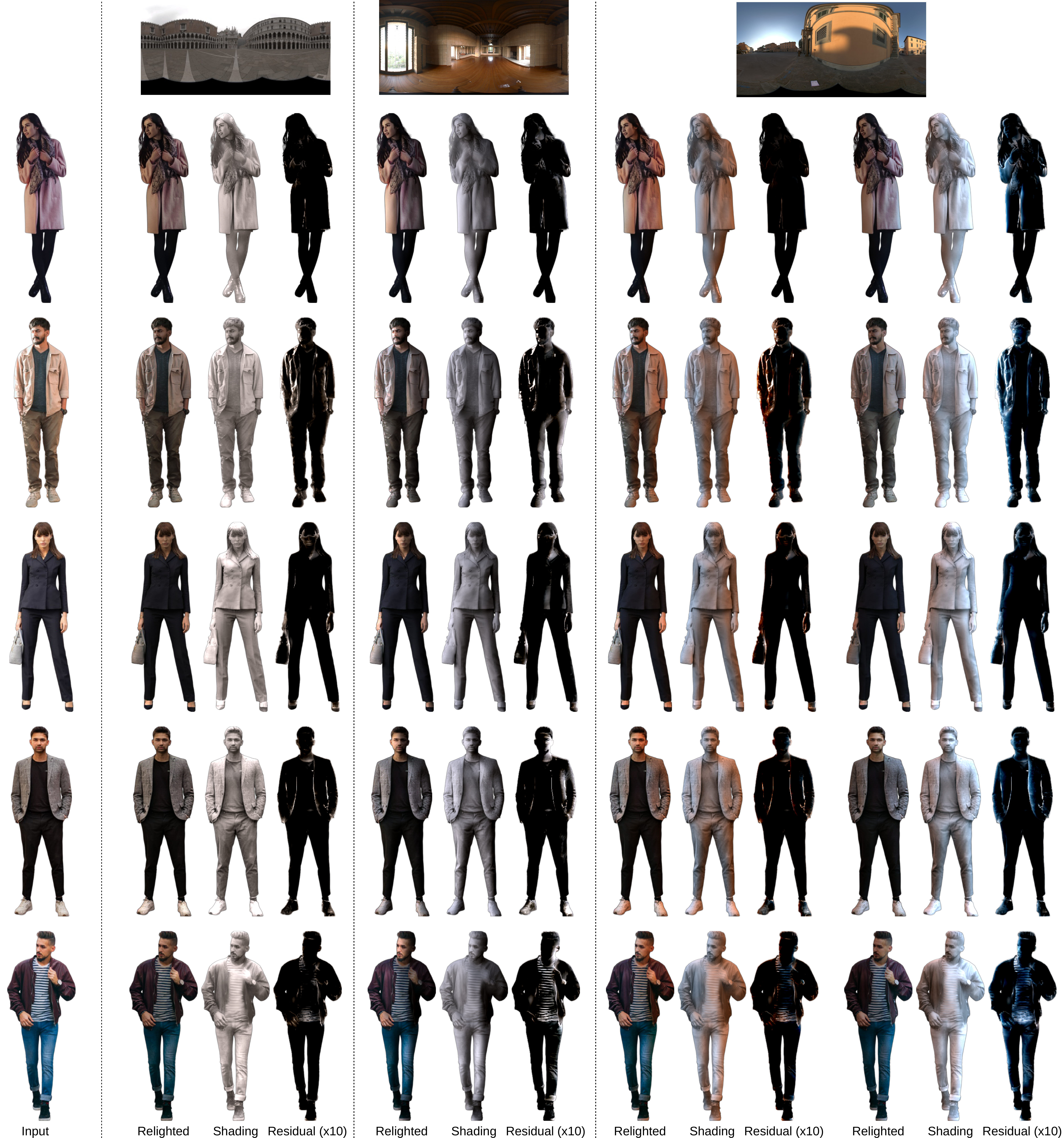}
    \caption{Relighting results for three different illuminations (\emph{doge}, \emph{ennis}, and \emph{pisa}) and five different input images. Last two columns feature the same illumination under two different rotations. In each case, we show the relighted image, and the reconstructed shading and residual terms. Our model is capable of producing a compelling relighting result for a varied set of input images and illuminations, including both indoors and outdoors cases. The residual term has been scaled by a factor of 10 for visualization purposes.
    \vspace{3em}}
    \label{fig:results_photos_relight}
\end{figure*}

Finally, in Figure~\ref{fig:results_photos_relight} we show a variety of relighting results under different illuminations (refer to the supplemental material for the full set, a table with quantitative metrics, as well as a video when rotating the illumination maps). For each input photo and illumination map we show the final relighted image, and the reconstructed shading and residual terms.

\subsection{Ablation Studies}
\label{subsec:ablation}

We evaluate the contribution of our design choices with a series of ablation experiments performed on both the synthetic images and the real photographs. 
In particular, we first compare the performance of our model (\emph{Ours}) without the residual generator predicting $\textbf{E}$ (\textit{Without $\textbf{E}$}) and without including the PRT decomposition in the architecture of the generators (\textit{Without PRT decomposition}). 
Then, we evaluate the impact of the logarithmic loss $\mathcal L_{\log}$ in the prediction of $\textbf{T}$ and $\textbf{L}$ (\textit{Without $\mathcal L_{\log}$}), as well as the performance of our model when using only nine coefficients (\emph{With $N=2$}). 
To avoid using a constant albedo in Equation~\ref{eq:render_prt_residual}, we combine the different terms that define reflectance (\emph{With $\textbf{T}^*$}) into a single vector $\textbf{T}^* = (\rho * \textbf{T} + \textbf{E})$.
Last, to showcase the benefit of our reflectance, we have trained a model using purely Lambertian materials in our data (\textit{Lambertian materials}).

Table~\ref{tab:ablation_real_and_synthetic} shows the results (including albedo and shading for synthetic images) for the L1, L2, and PSNR metrics for all the ablation studies.  All options yield significantly inferior results when compared with our full model. Figure~\ref{fig:ablation_real} further illustrates this on an example using a real photograph. One could think that the model \emph{With $\textbf{T}^*$} would obtain better performance since it does not need to assume a constant albedo $\rho$ in the reflectance. However, $\textbf{T}^*$ requires estimating 25 different RGB maps (with $N=4$), leading to additional complexity that hinders convergence and produces higher errors.

\section{Discussion}
\label{sec:discussion}

We have presented a model for human relighting that requires a single image as input. We lift the assumption on Lambertian materials and include a better approximation of material reflectance in our transport function. Moreover, we introduce an additional residual term which further mitigates errors in the PRT-based final reconstruction. 
This additional term becomes increasingly relevant for challenging illuminations, such as backlighting, where the overall dark appearance of the image does not allow for an accurate estimation of the PRT terms. The resulting errors are absorbed by our residual, helping to produce good final reconstructions.
Overall our results show compelling estimations of albedo and shading (transport and illumination), leading to accurate relighting reconstructions for both synthetic images and real photographs. 

\begin{figure}[!t]
    \centering
    \includegraphics[width=0.75\linewidth]{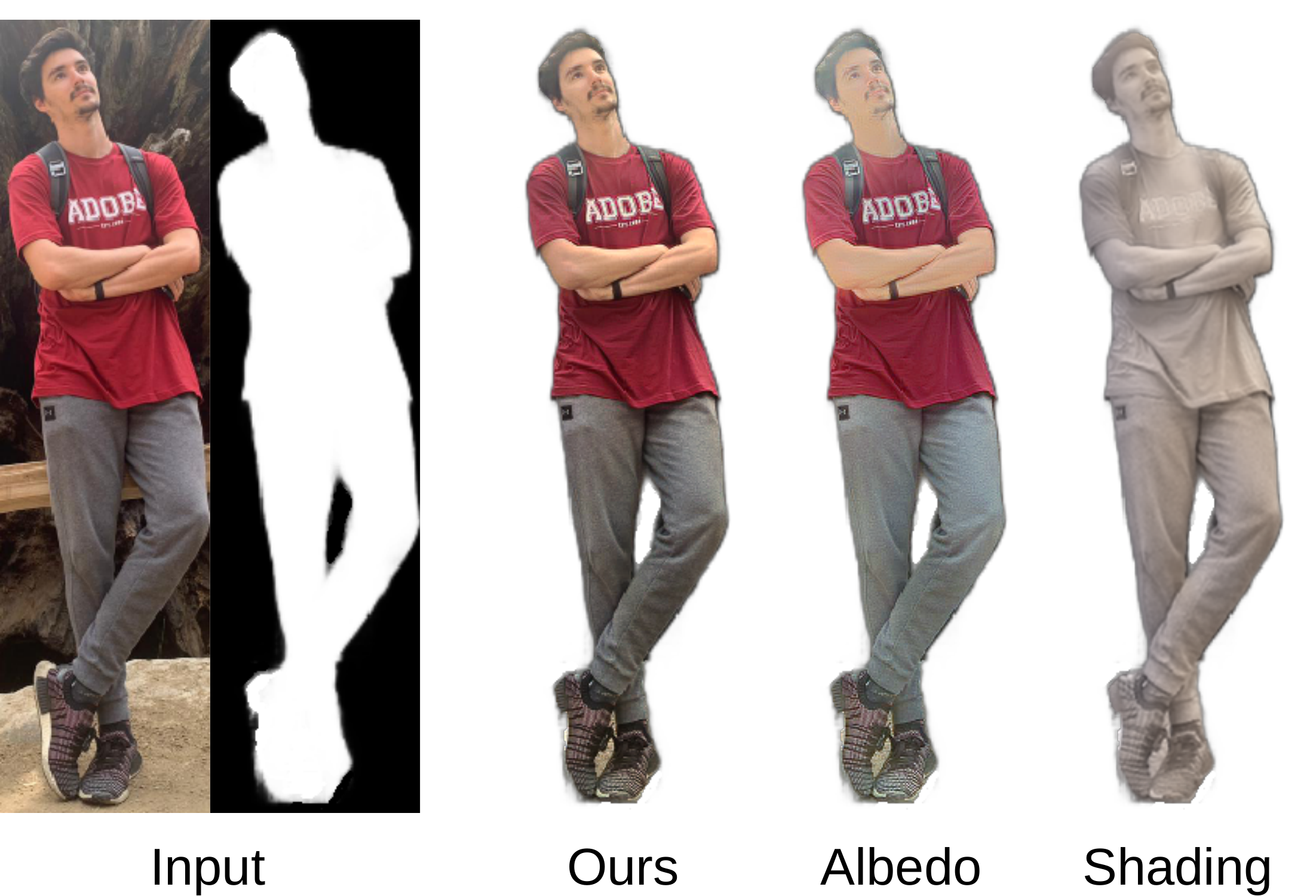}
    \caption{Example of the limitations of our model. The strong presence of stray light in the input image leads to an excessively flat albedo, seen especially in the head and shoulders area, while some texture details appear in the shading image.}
    \label{fig:limitations_photo}
\end{figure}

Nevertheless, our work is not free of limitations. Figure~\ref{fig:limitations_photo} shows a difficult case with a real photograph as input. While our reconstruction is still plausible, the strong presence of stray light (especially on top) leads to an excessively flat, milky estimation of the albedo in the head and shoulders area. 
Also, our shading reconstruction carries traces of texture details in the T-shirt, which remains an open problem in intrinsic images decomposition.

Human relighting poses many challenges not fully investigated in this work. Besides making the model more robust to poorly lit input images, being able to take into account other lighting effects such as subsurface scattering~\cite{jimenez2015separable}, anisotropy in cloth materials~\cite{aliaga2017appearance}, or more complex reflectance models, remain interesting open problems. 
Moreover, one implicit problem of SH-based lighting is the need for a large number of coefficients to reconstruct high-frequency details. While we mitigate this problem by introducing the residual term, complex high-frequency effects are still an open challenge.
Another exciting avenue of future work is to extend the potential of our approach, for instance by using contrastive loss functions, or proposing self-supervised schemes that would avoid having to generate additional synthetic data.
\section*{Acknowledgements}
    We want to thank the anonymous reviewers for their feedback on the manuscript; also, thanks to Ibon Guillen, and Adrian Jarabo for the occasional discussions about the paper. 
    This project has received funding from the European Research Council (ERC) under the European Union’s Horizon 2020 research and innovation programme (CHAMELEON project, grant agreement No 682080), from the European Union’s Horizon 2020 research and innovation programme under the Marie Sklodowska-Curie grant agreements No 765121 and 956585, from the Spanish Ministry of Economy and Competitiveness (project PID2019-105004GB-I00), and generous gifts from Adobe Systems.

\printbibliography                

\newpage

\end{document}